\documentclass[10pt,twocolumn,letterpaper]{article}

\usepackage[pagenumbers]{wacv} % To force page numbers, e.g. for an arXiv version

\usepackage{graphicx}
\usepackage{amsmath}
\usepackage{amssymb}
\usepackage{booktabs}

\usepackage{multirow}
\usepackage{makecell}

\usepackage[pagebackref,breaklinks,colorlinks]{hyperref}

\usepackage[capitalize]{cleveref}
\crefname{section}{Sec.}{Secs.}
\Crefname{section}{Section}{Sections}
\Crefname{table}{Table}{Tables}
\crefname{table}{Tab.}{Tabs.}

\def\wacvPaperID{951} % *** Enter the WACV Paper ID here
\def\confName{WACV}
\def\confYear{2025}

\newcommand{\name}{Semantic Prompt Learning for WSSS}
\newcommand{\namebf}{\textbf{Sem}antic \textbf{P}rompt \textbf{Le}arning for WSS\textbf{S}}
\newcommand{\nameshort}{SemPLeS}
\newcommand{\namea}{Segment-Label Matching}
\newcommand{\nameb}{Contrastive Prompt Learning}
\newcommand{\namec}{Prompt-guided Semantic Refinement}

\begin{document}

\title{Semantic Prompt Learning for Weakly-Supervised Semantic Segmentation}

\author{Ci-Siang Lin$^{1,2\text{*}}$ \qquad Chien-Yi Wang$^{2}$ \qquad Yu-Chiang Frank Wang$^{1, 2}$ \qquad Min-Hung Chen$^{2}$ \\
$^{1}$Graduate Institute of Communication Engineering, National Taiwan University, Taiwan \qquad $^{2}$NVIDIA\\
Project page: \url{https://projectdisr.github.io/semples/}
}
% \and

\let\oldtwocolumn\twocolumn
\renewcommand\twocolumn[1][]{%
    \oldtwocolumn[{#1}{
    \begin{center}
           \includegraphics[width=1.0\linewidth]{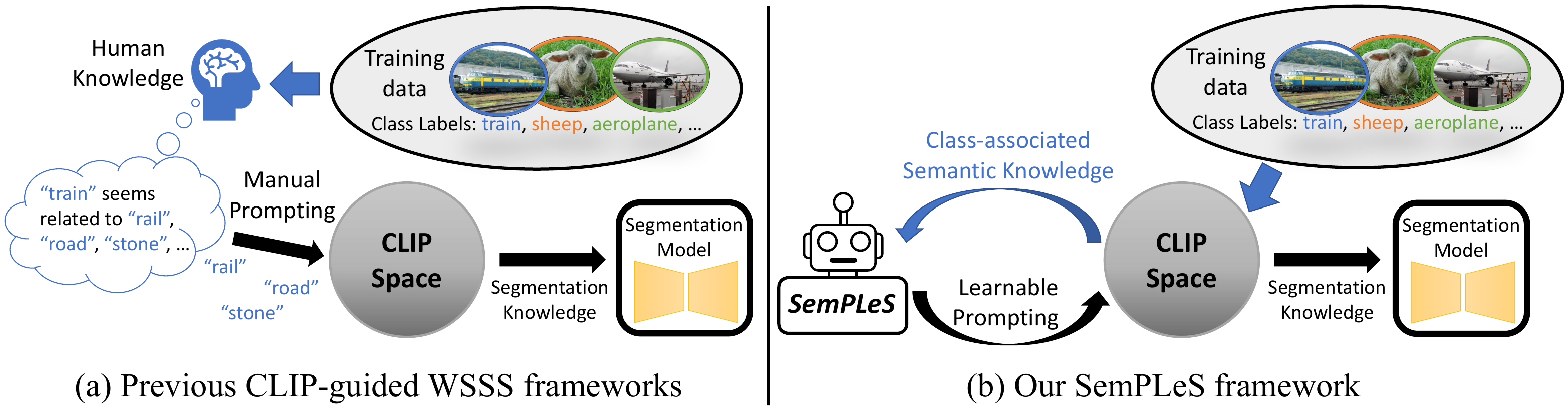}
           \vspace{-6mm}
           \captionof{figure}{(a) Previous CLIP-guided WSSS methods~\cite{xie2022clims,lin2023clip} rely on manual prompt engineering involved with heuristic human knowledge, while (b) our proposed \textbf{\textit{\nameshort}}~framework automatically discovers and learns prompts embedded with class-associated semantic knowledge from the CLIP latent space without any manual effort.}
           \label{figure:teaser}
           \vspace{-1mm}
        \end{center}
    }]
}
\maketitle

\def\thefootnote{*}\footnotetext{~Work done during an internship at NVIDIA.}

\begin{abstract}
\vspace{-3mm}
Weakly-Supervised Semantic Segmentation (WSSS) aims to train segmentation models using image data with only image-level supervision. Since precise pixel-level annotations are \textit{not} accessible, existing methods typically focus on producing pseudo masks for training segmentation models by refining CAM-like heatmaps. However, the produced heatmaps may capture only the discriminative image regions of object categories or the associated co-occurring backgrounds. To address the issues, we propose a \textit{\namebf~(\textbf{\nameshort})} framework, which learns to effectively prompt the CLIP latent space to enhance the semantic alignment between the segmented regions and the target object categories.
More specifically, we propose \textit{\nameb}~and \textit{\namec}~to learn the prompts that adequately describe and suppress the co-occurring backgrounds associated with each object category. In this way, \textit{\nameshort} can perform better semantic alignment between object regions and class labels, resulting in desired pseudo masks for training segmentation models. The proposed \textit{\nameshort} framework achieves competitive performance on standard WSSS benchmarks, PASCAL VOC 2012 and MS COCO 2014, and
shows compatibility with other WSSS methods. 
% Project page: \url{https://projectdisr.github.io/semples/}.

\end{abstract}
\vspace{-6mm}
\section{Introduction}
Semantic segmentation aims to classify every pixel in images to identify object categories and the associated regions, which can benefit various applications in the real world~\cite{ronneberger2015u, meyer2017improved, zendel2022unifying}.
While promising results have been presented by fully-supervised approaches~\cite{chen2015semantic, long2015fully, chen2017deeplab, chen2017rethinking, zhao2017pyramid, chen2018encoder, zhao2018psanet}, collecting pixel-level annotations could be time-consuming and expensive, and therefore limits the scalability and practicality of fully-supervised methods.
To address this issue, Weakly-Supervised Semantic Segmentation (WSSS) has emerged as an alternative approach to train segmentation models with only coarse or incomplete annotations such as bounding boxes~\cite{khoreva2017simple}, scribbles~\cite{lin2016scribblesup}, points~\cite{bearman2016s}, or image-level labels. Among these annotation forms, \textit{image-level labels} which indicate the presence or absence of certain object categories are commonly used due to the efficiency in data collection and the availability in various benchmark image datasets. Since precise annotations of object positions are \textit{not} observed, learning to localize and segment object categories from image-level supervision is particularly challenging. 
Most existing methods~\cite{chang2020weakly, wang2020self, ru2022learning, xu2022multi,jo2023mars,kwon2024learning} focus on producing pseudo ground truth masks by learning CAM-like heatmaps~\cite{zhou2016learning, selvaraju2017grad} with class labels as discriminative supervision.
Despite the shown efficacy, the learned CAMs may still miss relevant regions of target object categories and fail to cover the entire object. Furthermore, co-occurring backgrounds associated with certain object categories may also be falsely activated (\textit{e.g.}, rails in a photo of a train). Consequently, learning precise image regions that align with the semantics of target objects from weak supervision remains a challenging task.

With the rapid growth in the amount of image and text data in recent years, several vision-language models~\cite{lu2019vilbert,chen2020uniter,radford2021learning} have been proposed to bridge the underlying semantics between the two modalities. 
Given that both the images and the associated class labels (object names) are available in the setting of WSSS, the underlying image-text semantics from the CLIP~\cite{radford2021learning} latent space can be leveraged to enhance the quality of CAMs and pseudo masks.
Recent CLIP-based methods~\cite{xie2022clims,lin2023clip,yang2024foundation,murugesan2024prompting,xu2023mmcst} mainly focus on designing text prompts or prompt learning techniques for the text encoder. Despite the effectiveness demonstrated, they either consider only the foreground class prompts, or rely on general background prompts (\eg, ``a photo of rail'', ``a photo of road'', \etc) defined by additional manual efforts and heuristic human knowledge, as shown in Fig.~\ref{figure:teaser}~(a). Moreover, such manually-defined prompts may not fully exploit the knowledge in the CLIP latent space.

In this paper, we aim to fully exploit the CLIP latent space to benefit the weakly-supervised semantic segmentation problem without manual prompting. To achieve this goal, we propose a \textit{\namebf~(\textbf{\nameshort})} framework to learn prompts embedded with class-associated semantic knowledge \textit{discovered} from the CLIP latent space, as shown in Fig.~\ref{figure:teaser}~(b), where the learned prompts can enhance the semantic alignment between the segmented regions and the target object categories with image-level labels.
More specifically, we perform image-text contrastive learning under the guidance of CLIP and train a mask generator to generate class activation maps. Such produced object masks, however, might not be sufficiently precise, and the co-occurring backgrounds associated with the object categories may be falsely activated. To alleviate this problem, we uniquely present \textit{\nameb}~and \textit{\namec}~to suppress class-associated background regions. In \textit{\nameb}, we learn prompts to capture co-occurring backgrounds from images and class labels. Without manually defining the background texts, our learned prompts would properly describe the backgrounds associated with each object category. Under the guidance of our learned class-associated background prompts, we further suppress co-occurring backgrounds from the activation maps via \textit{\namec}. With the above-designed learning strategy, the semantic matching between object regions and the associated class labels will be enhanced, resulting in precise pseudo masks desired for training segmentation networks. The proposed \textit{\nameshort}~framework achieves competitive performance on standard WSSS benchmarks, PASCAL VOC 2012 and MS COCO 2014. Moreover, our proposed \textit{\nameshort}~framework can integrate and improve other WSSS methods including CNN-, Transformer-, and foundation model-based ones, confirming its effectiveness and compatibility. 

\vspace{2mm}
In summary, our contributions are three-fold:
\begin{itemize}

    \item We propose a novel \textit{\namebf~(\textbf{\nameshort})} framework, which fully exploits the CLIP latent space to benefit the weakly-supervised semantic segmentation without manual prompting. Additionally, our \textit{\nameshort}~framework shows compatibility with other WSSS methods, including CNN-, Transformer-, and foundation model-based ones.

    \item In \textit{SemPLeS}, we present \textit{\nameb} to learn prompts embedded with class-associated semantic knowledge. With no need to manually define background texts, our learned prompts would properly capture co-occurring backgrounds associated with distinct object categories.

    \item With the derived prompts, our \textit{\namec} learns to suppress co-occurring backgrounds while enhancing the semantic alignment between object regions and the associated class labels, resulting in precise pseudo masks and competitive segmentation performance in WSSS.

\end{itemize}

\section{Related Works}

\subsection{Weakly-Supervised Semantic Segmentation}

Existing WSSS approaches typically follow a three-stage learning process. Firstly, the image-level labels are utilized as supervision to generate Class Activation Maps (CAMs) \cite{zhou2016learning,selvaraju2017grad}. Secondly, the CAMs are refined by using dense CRF~\cite{krahenbuhl2011efficient} or pixel affinity-based methods~\cite{ahn2018learning,ahn2019weakly} to obtain pseudo masks. Lastly, the pseudo masks are further exploited to train segmentation networks. Among all the stages, producing precise CAMs (\ie, the first stage) is the main focus of WSSS, and various approaches have been proposed to improve the quality of CAMs~\cite{fan2020learning,chen2022self,chen2022class,xie2022c2am,du2022weakly,jiang2022l2g,li2022towards,lee2022weakly,yoon2022adversarial,wu2022adaptive,li2022expansion}. With the rapid development and the success of vision transformers~\cite{dosovitskiy2021image}, recent approaches~\cite{ru2022learning,xu2022multi,rossetti2022max,ru2023token,cheng2023out,xu2023mmcst,peng2023usage,zhu2023weaktr} generate finer activation maps based on the patch-level affinity learned from the attention layers. Very recently, several works~\cite{chen2023segment,jiang2023segment,sun2023alternative,chen2023weakly,yang2024foundation} exploit foundation segmentation models (\eg, SAM~\cite{kirillov2023segment}) to enhance the quality of the pseudo masks. On the other hand, there are also end-to-end WSSS works~\cite{ru2023token,wu2024masked} which do not require multiple training stages, yet their performances are inferior to standard 3-stage methods.
In general, most WSSS methods learn CAMs through object classification, overlooking the textual semantics of class labels. Instead, our method exploits vision-language models to discover class-associated semantic knowledge, therefore producing high-quality CAMs for segmentation.

\subsection{CLIP-based Semantic Segmentation}

Recently, the Contrastive Language-Image Pretraining (CLIP) model~\cite{radford2021learning} has been adopted in semantic segmentation tasks thanks to the generalized knowledge learned from a large corpus of image-text pairs.
Given the generalization capability, a number of zero-shot/open-vocabulary approaches~\cite{li2022language,luddecke2022image,ding2022decoupling,rao2022denseclip,xu2022simple,ghiasi2022scaling,liang2023open,xu2023side,xu2023open} exploit CLIP to segment the classes which are unseen during training. However, these methods still require mask annotations during training. To minimize the annotation effort, CLIP has also been adopted to improve unsupervised methods~\cite{zhou2022extract,shin2022reco,he2023clip}. Nevertheless, the segmentation performance is still unsatisfactory and is not desired for further applications.
On the other hand, CLIP has also been utilized to benefit WSSS~\cite{xie2022clims,lin2023clip,yang2024foundation,murugesan2024prompting,xu2023mmcst}. These works mainly focus on designing text prompts or prompt learning techniques for the text encoder. However, they either consider only the foreground class prompts, or rely on general background prompts defined by additional manual efforts and heuristic human knowledge. Moreover, such manually-defined prompts may not fully exploit the knowledge in the CLIP latent space.
In contrast, with no need for any manual efforts, our proposed \textit{\nameshort}~framework \textit{automatically} learns prompts embedded with class-associated semantic knowledge discovered from the CLIP latent space.

\subsection{Prompt Learning}

In natural language processing (NLP), prompting~\cite{liu2023pre} involves giving a text-based input such as a sentence or phrase to obtain desired responses from language models. Driven by the recent success of pre-trained vision-language models (\eg, CLIP~\cite{radford2021learning}),
there has been an increasing interest in identifying proper prompts for computer vision tasks. Early work relies on prompt engineering to identify text templates (\eg, ``a photo of {}'') describing classes of interest to obtain underlying knowledge.
However, such a trial and error approach generally takes a large amount of time and effort and also requires expertise about the task. To tackle the problem, prompt learning~\cite{zhou2022learning,zhou2022conditional,jia2022visual} is proposed to replace the manually-defined text templates with a set of learnable context vectors preceding the class names to automate the prompting process. 
Distinct from these prompt learning methods, our \textit{\nameshort}~framework aims to capture class-associated semantic knowledge for \textit{segmentation} purposes rather than replacing general text templates like ``a photo of'' in classification tasks. 

\section{Proposed Method}
\begin{figure*}[!t]
  \centering
  \includegraphics[width=\linewidth]{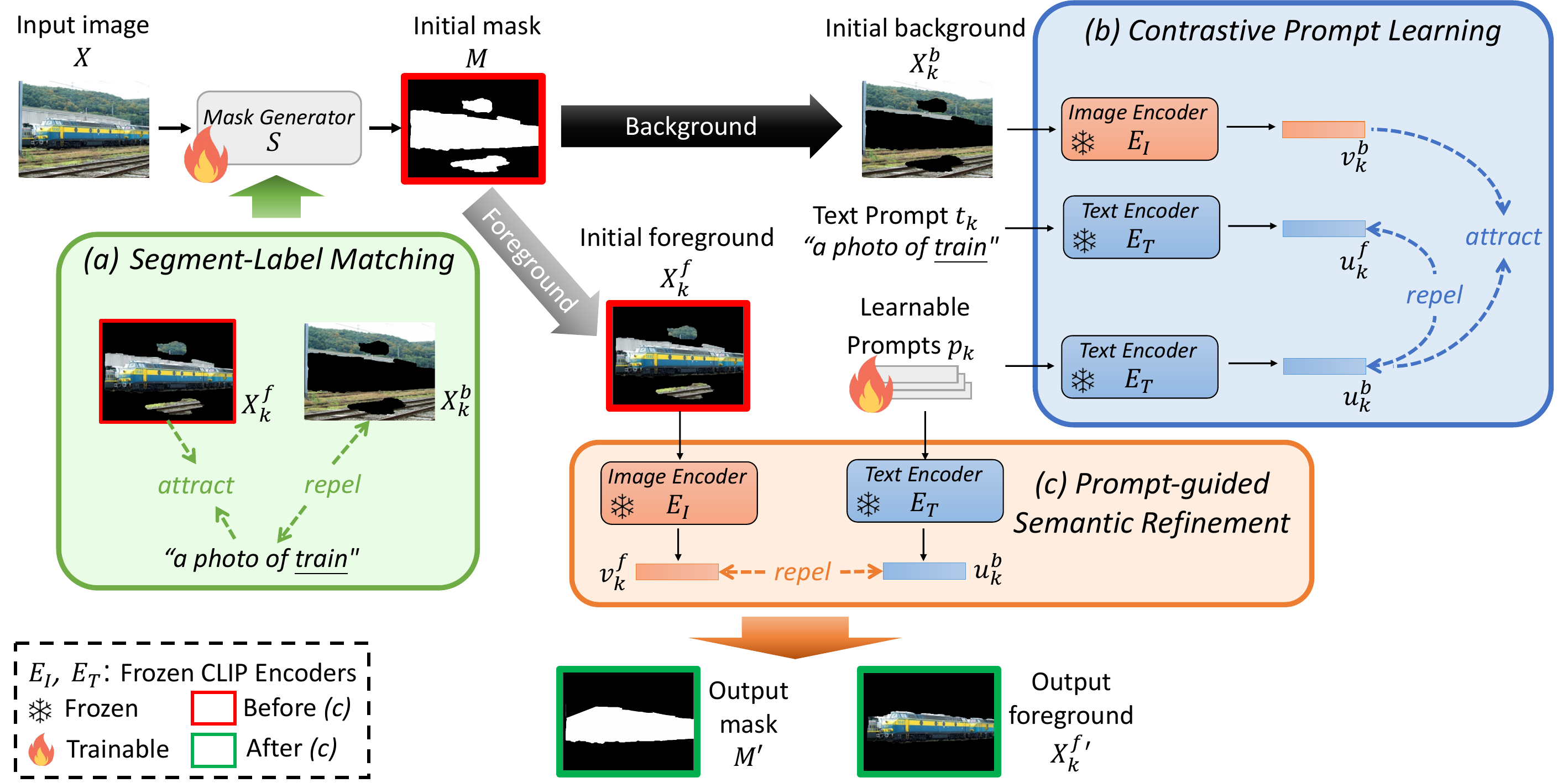}
  \caption{An overview of our proposed \textit{\textbf{\nameshort}} framework. We first introduce \textit{(a)~\namea}, which leverages image-text contrastive learning to train the mask generator $S$ and produce initial object masks $M$. Such derived masks are still coarse and may falsely include co-occurring backgrounds.
  To achieve class-associated mask refinement and produce the refined mask $M'$, we propose \textit{(b)~\nameb} to automatically learn prompts $p_k$ embedded with semantic knowledge from the CLIP latent space, followed by \textit{(c)~\namec} to suppress co-occurring backgrounds associated with each category $k$.
  }
  \label{figure:model}
  \vspace{-4mm}
\end{figure*}
\subsection{Problem Formulation and Model Overview} \label{sec:overview}

We first define the problem setting and notations used in this paper. In weakly-supervised semantic segmentation (WSSS), we assume that there is a set of $N$ images $X$ with the associated image-level labels $y$, where $X \in \mathbb{R}^{H \times W \times 3}$ and $y \in \{0, 1\}^K$ is a multi-hot vector indicating the presence or absence of $K$ object categories. Without access to pixel-wise annotations, 
we propose a novel \textit{\namebf~(\textbf{\nameshort})}~framework to exploit CLIP~\cite{radford2021learning} to learn prompts that can enhance the semantic alignment between the segmented regions and the target object categories. 

As shown in Fig.~\ref{figure:model}, we first introduce \textit{(a)~\namea} in our \textit{\nameshort}~framework, which leverages image-text contrastive learning to produce initial object masks $M$ from our mask generator $S$.
To suppress falsely activated backgrounds in such masks (\textit{e.g.}, $X^f_k$ in the red box), we uniquely propose~\textit{(b)~\nameb} and \textit{(c)~\namec}. The former learns class-associated prompts $p_k$ to capture co-occurring backgrounds from images and labels, while the latter takes the derived prompts to disregard co-occurring backgrounds from the object masks (\textit{e.g.}, $X^f_k{'}$ in the green box). By jointly enforcing vision-language matching and suppression objectives, our framework would enhance the semantic alignment between object regions and the associated text labels, resulting in precise segmentation results.

\subsection{\name}

\subsubsection{\namea} \label{sec:SLM}
Given an input image $X$, our mask generator $S$ is designed to produce foreground masks $M=S(X)$ for target object categories. Since pixel-wise annotations are not available, we choose to leverage vision-language models to guide the learning of our mask generators from image-level supervision. To be more precise, we exploit the joint latent space for images and texts from CLIP to learn object regions aligned with the associated text labels. To achieve this, an image-text triplet (\textit{i.e.}, foreground-background-text) would be formulated to perform contrastive learning, as illustrated in Fig.~\ref{figure:model} (a). For the $k$th ground truth category which presents in the input image $X$ (\textit{i.e.}, $y_k=1$), we derive the foreground image $X^f_k = M_k \cdot X$ by applying the $k$th predicted mask $M_k$ to the original image $X$. Similarly, we reverse the predicted mask to obtain the background regions $X^b_k = (1-M_k) \cdot X$. As for the text input $t_k$, we adopt the common prompt template ``a photo of \{\}'' filled with the $k$th class name to describe the category of interest. With the triplet [$X^f_k$, $X^b_k$, $t_k$] serving as the input of the image encoder $E_I$ and text encoder $E_T$ pre-trained by CLIP, we perform image-text contrastive learning to maximize the cosine similarity between $X^f_k$ and $t_k$ for the foreground, while the similarity of $X^b_k$ and $t_k$ would be minimized to repel the background. Therefore, our matching loss $L_{match}$ would be formulated as follows:
\begin{equation}
\begin{aligned}
\label{eq:loss_contra}
L_{match} = \quad
&\mathbb{E}_{X}\left[-log ( sim(v^f_k, u^f_k))\right] + \\
& \mathbb{E}_{X}\left[-\lambda_{b} \cdot log ( 1- sim(v^b_k, u^f_k))\right], \\
\textit{where} \,\, v^f_k = E_I&(X^f_k), \,\, v^b_k = E_I(X^b_k), \,\, u^f_k = E_T(t_k).
\end{aligned}
\end{equation}

Here, $\lambda_b$ is the loss weight for repelling backgrounds and $sim$ refers to cosine similarity. 
Note that we keep the image encoder $E_I$ and the text encoder $E_T$ frozen during training and preserve the latent space learned from CLIP to avoid potential overfitting. With the above \textit{\namea}, our mask generator $S$ is encouraged to distinguish foregrounds and backgrounds with the associated text labels. However, as noted above, such masks learned from image-level supervision are still coarse, and may falsely include co-occurring backgrounds associated with certain object categories. Therefore, the above image-text matching is not sufficient to achieve precise segmentation.

\vspace{-2mm}

\subsubsection{\nameb} \label{sec:SPL}
To address the coarse mask issues, the previous language-guided approach~\cite{xie2022clims} exploits vision-language models to refine the masks with manual prompting techniques. However, these methods require additional prompt engineering efforts with human knowledge involved. Moreover, manual prompting may not be able to fully exploit vision-language representation space. To tackle these problems, we propose \textit{\nameb} (Fig.~\ref{figure:model} (b)) to learn prompts embedded with semantic knowledge from vision-language models, facilitating the following object mask refinement. 
Different from the previous work, we employ a sequence of learnable prompts $p_k$ as the input of the text encoder $E_T$ to describe backgrounds for each distinct category $k$. 
Specifically, to align the prompts $p_k$ with the background image $X^b_k$, we maximize the similarity of their representations in the latent space via $L^I_{prompt}$. On the other hand, to avoid describing the target object category, we encourage the feature similarity between $p_k$ and $t_k$ to be low with $L^T_{prompt}$. 
Thus, the prompt learning loss $L_{prompt}$ is defined as below:
\begin{equation}
\begin{aligned}
\label{eq:loss_prompt}
L_{prompt} = \quad &L^I_{prompt} + \lambda_T \cdot  L^T_{prompt}\\
 = \quad &\mathbb{E}_{X}\left[-log ( sim(u^b_k, v^b_k))\right] + \\
&\mathbb{E}_{X}\left[-\lambda_T \cdot log(1-sim(u^b_k, u^f_k))\right],\\
\textit{where} \,\, u^b_k = E_T&(p_k), \,\, v^b_k = E_I(X^b_k), \,\, u^f_k = E_T(t_k).
\end{aligned}
\end{equation}

Here, the mask generator $S$ is fixed and $p_k$ is the only trainable part for loss $L_{prompt}$, and $\lambda_{T}$ is the loss weight for minimizing the similarities to the text labels. Once the above learning is complete, our prompts $p_k$ would represent co-occurring backgrounds for each category $k$ without requiring manually-defined background prompts, and is therefore preferable to existing CLIP-based methods~\cite{xie2022clims,lin2023clip,yang2024foundation,murugesan2024prompting,xu2023mmcst}. In addition, our \textit{\nameb}~aims to capture class-associated backgrounds for segmentation purposes, rather than replacing general text templates like ``a photo of \{\}'' for classification tasks as previous prompt learning methods~\cite{zhou2022learning,zhou2022conditional,jia2022visual} do.

\vspace{-1mm}

\subsubsection{\namec} \label{sec:PSR}
Finally, to suppress co-occurring background regions from the object mask $M$, our \textit{\nameshort} framework exploits the previously derived background prompts $p_k$ to perform \textit{\namec} (Fig.~\ref{figure:model} (c)). More specifically, we encourage our mask generator $S$ to produce refined masks $M'$ by excluding the semantic knowledge embedded in the background prompts $p_k$, while the objectives introduced in Eq.~(\ref{eq:loss_contra}) are retained to match the class labels. Hence, the refinement loss $L_{refine}$ and the total loss function $L_{total}$ are defined as follows:
\begin{equation}
\begin{aligned}
\label{eq:loss_total}
&L_{total} = L_{match} + \lambda \cdot L_{refine},\\ 
\textit{where} \quad & L_{refine} = \mathbb{E}_{X}\left[-log ( 1-sim(v^f_k, u^b_k)) \right].
\end{aligned}
\vspace{-2mm}
\end{equation}

Here, $\lambda$ is the weight for the refinement loss. It can be seen that, with the derived background prompts $p_k$ (fixed here) and the introduced refinement loss $L_{refine}$, the class-associated background regions would be suppressed from the foreground mask $M$, preventing possible false activation. More importantly, by jointly applying the matching and refinement objectives with image-level supervision, our \textit{\nameshort} framework advances vision-language learning to enhance the semantic alignment between the segmented regions and the target object categories, resulting in compact and complete object masks $M'$ desired for WSSS. It is worth noting that, the CLIP model and the learned prompts $p_k$ are leveraged to guide the learning of the mask generator $S$ in our framework, and hence only the mask generator $S$ is needed for producing object masks $M'$ in the WSSS pipeline when the training is complete.

\section{Experiments} \label{sec:experiments}

\subsection{Datasets and Evaluation Metrics}  \label{sec:dataset}

We train and validate our proposed framework on the benchmark semantic segmentation datasets, PASCAL VOC 2012~\cite{everingham2010pascal} and MS COCO 2014~\cite{lin2014microsoft}. The PASCAL VOC 2012 dataset contains $20$ object categories along with a background category. The original training, validation, and testing set consists of $1464$, $1449$, and $1456$ images, respectively. Following the common protocol in previous WSSS works, we use an augmented set of $10582$ images for training. The testing set results are obtained from the official evaluation website. As for the MS COCO 2014 dataset, the training and validation set contains $82081$ and $40137$ images from $80$ object categories, respectively.
The mean Intersection over Union (mIoU) is used as the evaluation metric for all experiments.

\subsection{Implementation Details} \label{sec:imple}

For CLIP~\cite{radford2021learning}, we use ViT-B/32~\cite{dosovitskiy2021image} as the image encoder. Following~\cite{xie2022clims}, we adopt the cosine feature similarity where non-positive scores are clamped to a small positive number. The learnable prompts are randomly initialized with the sequence length $K=30$. The default batch size is $64$. We set the initial learning rate to 5e-4 and 5e-6 and train our framework for $60$ epochs on PASCAL VOC 2012 and MS COCO 2014, respectively. For loss weights, we set $\lambda_b$, $\lambda_T$ and $\lambda$ as $2.4$, $0.02$ and $0.05$ for PASCAL VOC 2012 and $0.75$, $0.01$ and $0.2$ for MS COCO 2014.  The AdamW optimizer is adopted with the cosine scheduler.   
The proposed framework is implemented in PyTorch and trained with NVIDIA V100 GPUs. 

\begin{table}[t]
  \centering
\center
\resizebox{0.8\linewidth}{!}{
\begin{tabular}{lc@{\hskip 0.2in}c}
     \Xhline{1pt}&&\\
     [-0.95em]
    Method & CAM & Mask  \\
    \hline &&\\[-0.8em]
    $\text{MCTformer}_\text{~~CVPR'22}$~\cite{xu2022multi} & 61.7 & 69.1 \\
    $\text{CLIMS}_\text{~~CVPR'22}$~\cite{xie2022clims} & 57.5 & 72.8 \\
    $\text{WeakTr}_\text{~~arXiv'23}$~\cite{zhu2023weaktr} & 65.9 & 74.2 \\
    $\text{CLIP-ES}_\text{~~CVPR'23}$~\cite{lin2023clip} & 58.6 & 75.0 \\
    $\text{FPR}_\text{~~ICCV'23}$~\cite{chen2023fpr} &  63.8 & 66.4\\
    $\text{D2CAM}_\text{~~ICCV'23}$~\cite{wang2023treating} & 58.0 & 71.4 \\
    $\text{USAGE}_\text{~~ICCV'23}$~\cite{peng2023usage} & 67.7 & 72.8 \\
    $\text{MCC}_\text{~~WACV'24}$~\cite{wu2024masked} & - & 73.0 \\
    $\text{POLE}_\text{~~WACV'24}$~\cite{murugesan2024prompting} & 59.0 & 74.2 \\
    $\text{DuPL}_\text{~~CVPR'24}$~\cite{wu2024dupl} & - & 76.0 \\
    \hline &&\\[-0.8em]
    \textbf{\nameshort~(Ours)} & \textbf{68.7} & \textbf{78.4} \\
    \Xhline{1pt}
    \end{tabular}
}
\vspace{-1mm}
  \caption{Quantitative results of CAMs (CAM) and the resulting pseudo masks (Mask) on PASCAL VOC 2012 \textit{train} set.
}
\label{tab:train}
  \vspace{-3mm}

\end{table}

\subsection{Quantitative Comparisons}
\label{sec:quantitative}
\begin{figure*}[!t]
  \centering
  \includegraphics[width=0.9\linewidth]{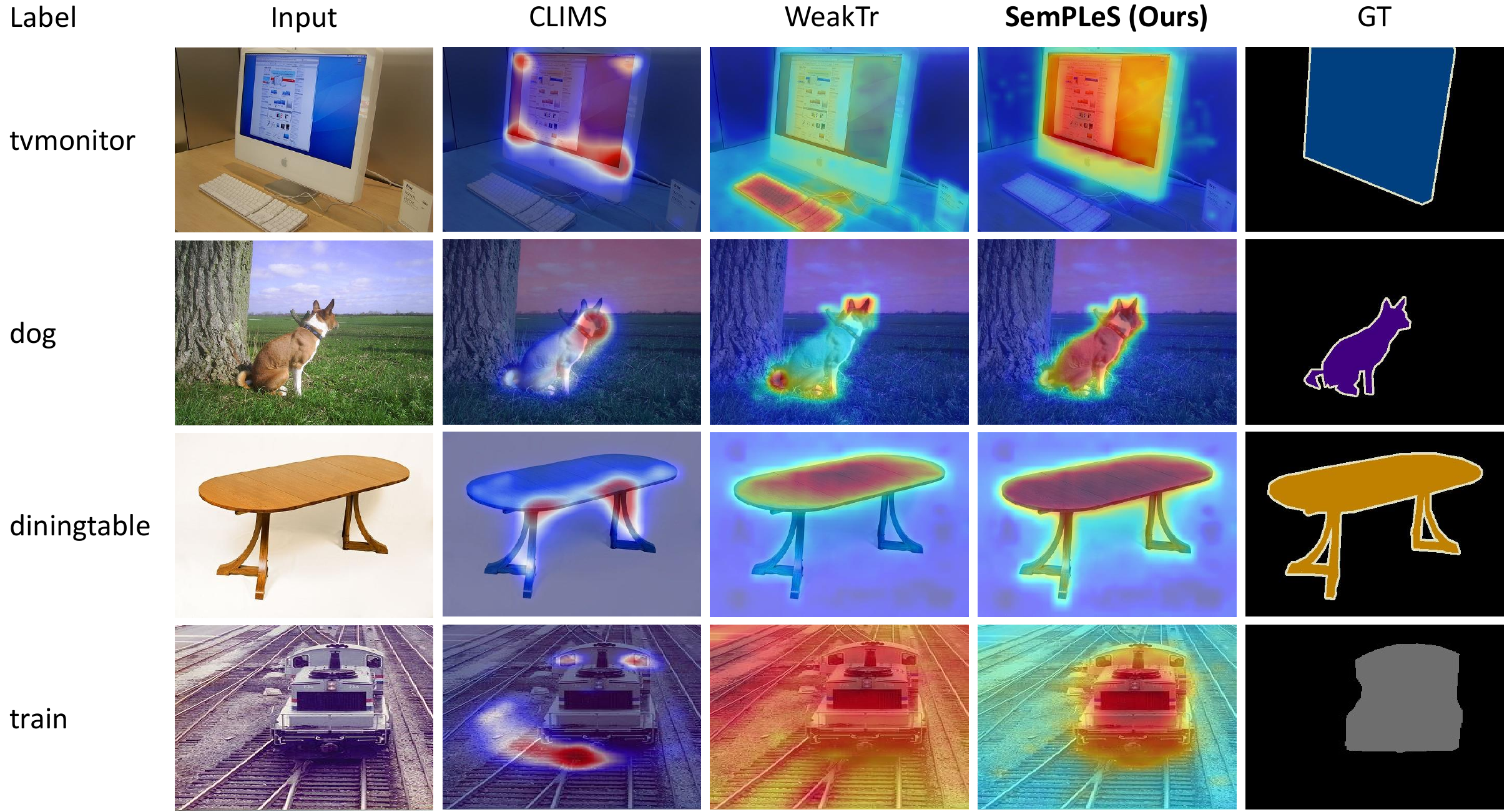}
  \vspace{-2mm}
  \caption{Qualitative results of CAMs. ``GT'' denotes the ground truth masks. We see that our proposed \nameshort~framework produces precise CAMs better aligned with the ground truth masks. 
  }
  \label{figure:pascal}
  \vspace{-3mm}
\end{figure*}
\begin{table}[!t]
  
  \centering
  \begin{tabular}{@{}l l c c @{}}
    \toprule
    Method & Backbone & \textit{val} & \textit{test} \\
				
    \midrule
    \multicolumn{4}{l}{\textit{\textbf{CNN-/Transformer-based approaches}}}\\
    SIPE$_\text{~~CVPR'22}$~\cite{chen2022self} & DL2-Res101 & 68.8 & 69.7 \\
    $\text{CLIMS}_\text{~~CVPR'22}$~\cite{xie2022clims} & DL2-Res101 & 70.4 & 70.0 \\
    $\text{MCTformer}_\text{~~CVPR'22}$~\cite{xu2022multi} & DL1-WRes38 & 71.9 & 71.6 \\
    $\text{CLIP-ES}_\text{~~CVPR'23}$~\cite{lin2023clip} & DL2-Res101 & 71.1 & 71.4 \\
    $\text{MMCST}_\text{~~CVPR'23}$~\cite{xu2023mmcst} & DL1-WRes38 & 72.2 & 72.2 \\
    $\text{LPCAM}_\text{~~CVPR'23}$~\cite{chen2023extracting} & DL1-WRes38 & 72.6 & 72.4 \\   $\text{FPR}_\text{~~ICCV'23}$~\cite{chen2023fpr} & DL2-Res101 & 70.3 & 70.1 \\
    $\text{D2CAM}_\text{~~ICCV'23}$~\cite{wang2023treating} & DL2-Res101 & 71.2 & 70.7 \\
    $\text{MCC}_\text{~~WACV'24}$~\cite{wu2024masked} & DeiT-B & 70.3 & 71.2 \\
    $\text{POLE}_\text{~~WACV'24}$~\cite{murugesan2024prompting} & DL2-Res101 & 71.5 & 71.4 \\
    $\text{SFC}_\text{~~AAAI'24}$~\cite{zhao2024sfc} & DL2-Res101 & 71.2 & 72.5 \\
    $\text{DuPL}_\text{~~CVPR'24}$~\cite{wu2024dupl} & ViT-B & 73.3 & 72.8 \\
    \midrule
    \multicolumn{4}{l}{\textit{\textbf{SAM-based approaches}}}\\
    $\text{SEPL}_\text{~~NeurIPSW'23}$~\cite{chen2023segment} & DL2-Res101 & 71.1 & -\\
    $\text{SG-WSSS}_\text{~~arXiv'23}$~\cite{jiang2023segment} & DL2-Res101 & 71.1 & 72.2 \\
    $\text{FMA-WSSS}_\text{~~WACV'24}$~\cite{yang2024foundation} & M2F-Swin-L & 82.6 & 81.6 \\
    \midrule
    \midrule
    \textbf{\nameshort~(Ours)} & DL2-Res101 & 73.9 & 73.8 \\
    \textbf{\nameshort~(Ours)} & M2F-Swin-L & \textbf{83.4} & \textbf{82.9} \\
    \bottomrule
  \end{tabular}
  \vspace{-2mm}
    \caption{Quantitative results of segmentation masks on PASCAL VOC 2012~\cite{everingham2010pascal} \textit{val} and \textit{test} sets. 
    ``Backbone'' denotes the segmentation network. ``DL'', ``Res'', ``WRes'', and ``M2F'' denote DeepLab~\cite{chen2017deeplab}, ResNet~\cite{he2016deep}, WideResNet~\cite{zagoruyko2016wide}, and Mask2Former~\cite{cheng2022masked}, respectively.
  }
  \label{tab:pascal_coco}
  \vspace{-5mm}
\end{table}

\begin{table}[!t]
  
  \centering
  \begin{tabular}{@{}l c @{}}
    \toprule
    Method & COCO \textit{val} \\				
    \midrule
    $\text{MCC}_\text{~~WACV'24}$~\cite{wu2024masked} & 42.3\\
    $\text{USAGE}_\text{~~ICCV'23}$~\cite{peng2023usage}& 42.7\\
    $\text{LPCAM}_\text{~~CVPR'23}$~\cite{chen2023extracting} & 42.8\\                
    $\text{FPR}_\text{~~ICCV'23}$~\cite{chen2023fpr} & 43.9\\
    $\text{D2CAM}_\text{~~ICCV'23}$~\cite{wang2023treating} & 44.0\\
    $\text{WeakTr}_\text{~~arXiv'23}$~\cite{zhu2023weaktr} & 44.4\\
    $\text{DuPL}_\text{~~CVPR'24}$~\cite{wu2024dupl} & 44.6 \\
    $\text{G-RAM-SAM}_\text{~~arXiv'23}$~\cite{chen2023weakly} & 54.6 \\
    $\text{FMA-WSSS}_\text{~~WACV'24}$~\cite{yang2024foundation} & 55.4 \\
    \midrule
    \textbf{\nameshort~(Ours)} & \textbf{56.1} \\
    \bottomrule
  \end{tabular}
  \vspace{-1mm}
  \caption{
  Quantitative results of the segmentation masks on MS~COCO~2014~\cite{lin2014microsoft} \textit{val} set.
  }
  \label{tab:coco}
  \vspace{-3mm}
\end{table}

To evaluate our proposed \nameshort~framework, we follow the standard WSSS pipeline and take our refined masks $M'$ as CAMs to produce pseudo masks. In Table~\ref{tab:train}, we compare the quality of CAMs and also the resulting pseudo masks with previous works. From the results in this table, we see that our \nameshort~achieves the best performance compared with previous weakly-supervised segmentation methods. Specifically, our CAMs achieve $\textbf{68.7\%}$ and the produced pseudo masks report $\textbf{78.4\%}$ in mIoU.
This verifies that, by exploiting CLIP to perform vision-language learning plus the designed prompt learning, our proposed \nameshort~framework successfully generates pixel-wise predictions from image-level supervision, which helps learn the following segmentation network. 

In Table~\ref{tab:pascal_coco}, by taking the derived pseudo masks to train the segmentation networks, we see that our \nameshort~achieves the best performance on PASCAL VOC and reports $83.4\%$ and $82.9\%$ mIoU on the validation and testing sets, respectively. Our method outperforms the previous work, FMA-WSSS~\cite{yang2024foundation}, by $0.8\%$ and $1.3\%$ mIoU on the validation and testing sets, respectively. 
In addition, our \nameshort~achieves the competitive performance of $56.1\%$ mIoU on MS COCO in Table~\ref{tab:coco}. 
The above results verify that our method is effective in performing semantic segmentation from image-level supervision. 

\vspace{-2mm}
\begin{figure*}[!t]
  \centering
  \includegraphics[width=0.95\linewidth]{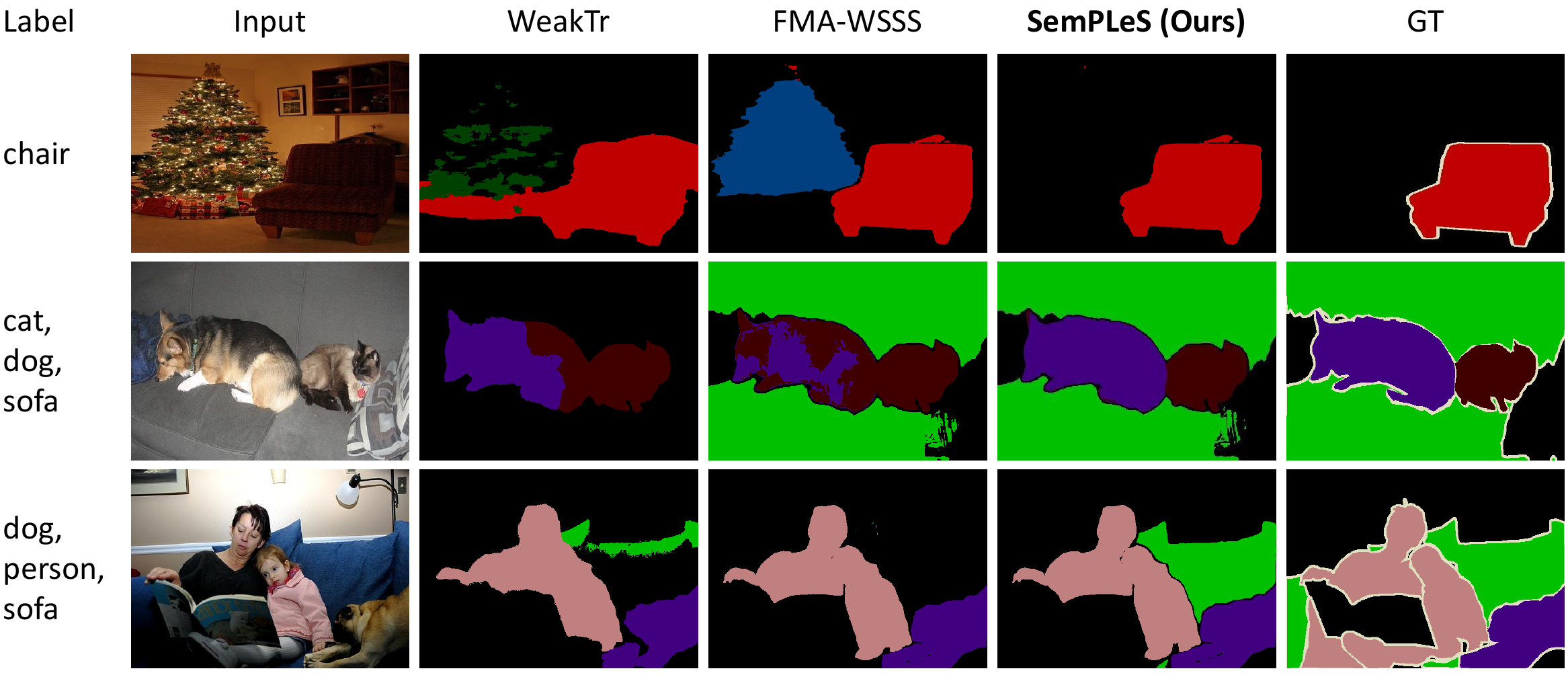}
  \vspace{-1mm}
  \caption{Qualitative results of segmentation maps. ``GT'' denotes the ground truth masks. 
  }
  \label{figure:seg}
    \vspace{-3mm}
\end{figure*}
\begin{table}[!t]
	\centering
        
		\begin{tabular}{l c c}
			\toprule
			Method & \textit{val} & \textit{test}\\
			\midrule
			\text{CLIMS}~\cite{xie2022clims}   & 70.4 & 70.0 \\
                \textbf{+\nameshort~(Ours)}  & \textbf{71.0} & \textbf{71.5}\\ 
                \midrule
                \text{WeakTr}~\cite{zhu2023weaktr} & 73.2 & 74.0 \\     
                \textbf{+\nameshort~(Ours)} & \textbf{74.2} & \textbf{74.8} \\
                \midrule
                \text{FMA-WSSS}~\cite{yang2024foundation} & 82.6 & 81.6 \\
                \textbf{+\nameshort~(Ours)} & \textbf{83.4} & \textbf{82.9} \\
			\bottomrule
	    \end{tabular}
\vspace{-1mm}
\caption{Quantitative results of our proposed \nameshort~framework based on different WSSS methods, including CNN- (CLIMS), Transformer- (WeakTr), and SAM-based (FMA-WSSS) ones.
    }
\label{tab:pascal_backbone}
\end{table}

\paragraph{Comparison with CLIP-based methods:}
In light of the constraint in WSSS (\ie, training using only class labels), several approaches have leveraged CLIP to enhance the quality of the produced CAMs by prompting, such as CLIMS~\cite{xie2022clims}, CLIP-ES~\cite{lin2023clip}, MMCST~\cite{xu2023mmcst}, and POLE~\cite{murugesan2024prompting}. 
However, they either consider only the foreground class prompts, or rely on general background prompts defined by additional manual efforts and human knowledge. Moreover, such manually-defined prompts may not fully exploit the knowledge learned in CLIP.
In contrast, our method automatically learns prompts embedded with class-associated semantic knowledge from the CLIP latent space with no need of any manual efforts, resulting in better performance than these CLIP-based methods in Table~\ref{tab:pascal_coco}.
\vspace{-3mm}
\paragraph{Comparison with SAM-based methods:}
Recently, SAM has been proposed to produce high-quality class-agnostic masks with overwhelming generalizability. Several approaches~\cite{chen2023segment,sun2023alternative,chen2023weakly,jiang2023segment} have explored the potential of leveraging SAM for WSSS, and most of them require additional foundation models (\eg, BLIP-2~\cite{li2023blip}, Grounding-DINO~\cite{liu2023grounding}, and RAM~\cite{zhang2023recognize}) to incorporate semantic information for semantic segmentation. 
Specifically, FMA-WSSS~\cite{yang2024foundation} exploits CLIP and achieves competitive performance among the above methods, and we further outperform FMA-WSSS with the proposed \nameshort~framework, as shown in Table~\ref{tab:pascal_coco}.

\vspace{-1mm}
\paragraph{Compatibility with other WSSS methods:}
We evaluate the compatibility of our method by integrating the proposed \nameshort~framework with other WSSS methods, including CNN-based~\cite{xie2022clims}, Transformer-based~\cite{zhu2023weaktr}, and SAM-based ones~\cite{yang2024foundation}.
The quantitative results are presented in Table~\ref{tab:pascal_backbone}. From this table, we see that our \nameshort~improves all types of methods, demonstrating the compatibility and robustness of the proposed framework. It is worth noting that, even though CLIMS~\cite{xie2022clims} already uses manually-defined prompts, our \nameshort~could still achieve further improvement with our learnable prompts. This shows that manually-defined prompts are limited and may not fully exploit the CLIP latent space, while our learnable prompts is able to automatically capture the semantic knowledge associated with target object categories to complement such pre-defined prompts, verifying the effectiveness of our designed prompt learning and proposed \nameshort~framework.

\vspace{1mm}

\subsection{Qualitative Comparisons}

For qualitative comparisons, our method shows more precise activation maps on various object categories and performs favorably compared with previous works, as shown in Fig.~\ref{figure:pascal}. In Fig.~\ref{figure:seg}, we also show that our segmentation results are superior to other WSSS methods. This validates that, by advancing image-text contrastive learning with learnable prompts, our \nameshort~would enhance the alignment between the segment regions and the target object categories, resulting in precise CAMs and segmentation masks better aligned with the ground truth.

\begin{figure}[!t]
  \centering
      \includegraphics[width=1.0\linewidth]{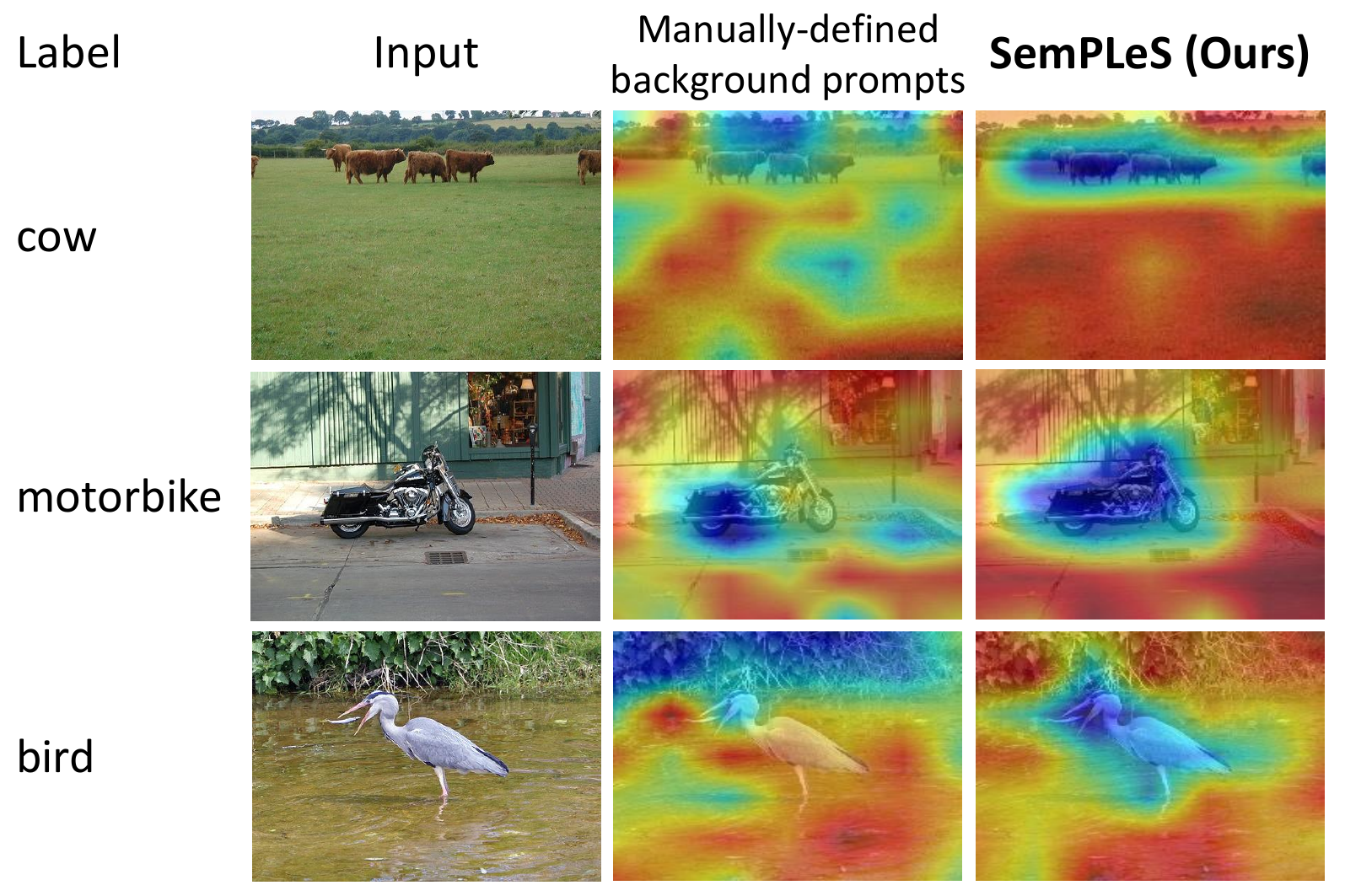}
  \vspace{-3mm}
  \caption{Visualization of the manually-defined background prompts~\cite{xie2022clims} and our learned prompts.}
  \label{figure:prompt}
\end{figure}

In addition, we also visualize the corresponding regions of our learned prompts by calculating the similarities to image patches with the text and image encoders of CLIP. As shown in Fig.~\ref{figure:prompt}, the manually-defined background prompts~\cite{xie2022clims} may falsely highlight the foreground objects (\eg, the bird example in the third row) due to their high co-occurrence when pre-training CLIP. Also, such manual prompts are limited and may fail to cover the whole background in images (\eg, the cow example in the first row).
In contrast, our learned prompts highlight all the background regions associated with each object category, showing the effectiveness of our \nameb. 

\begin{table}[!t]
	\centering
	\resizebox{1.0\linewidth}{!}{
		\begin{tabular}{c c c c c}
			\toprule
			$L_{match}$ & $L^T_{prompt}$ & $L^I_{prompt}$ & $L_{refine}$ & mIoU \\
			\midrule
			\checkmark &            &             &            & 67.6 \\
                \checkmark &            &             &       \checkmark     & 67.6 \\
                \checkmark & \checkmark &             &   \checkmark & 67.7 \\
			\checkmark &  & \checkmark  & \checkmark & 67.9 \\
			\checkmark & \checkmark & \checkmark  & \checkmark & \textbf{68.7} \\
			
			\bottomrule
	    \end{tabular}
 }
 \vspace{-1mm}
 \caption{Quantitative ablation studies of our loss functions. With both $L^T_{prompt}$ and $L^I_{prompt}$ applied (Eq.~(\ref{eq:loss_prompt})), the derived prompts would be desired to guide the semantic refinement through loss $L_{refine}$, resulting in the best performance.}
	\label{tab:loss}
\vspace{-4mm}
\end{table}

\vspace{2mm}

\subsection{Ablation Studies}

\begin{figure}[!t]
  \centering
  \includegraphics[width=1.0\linewidth]{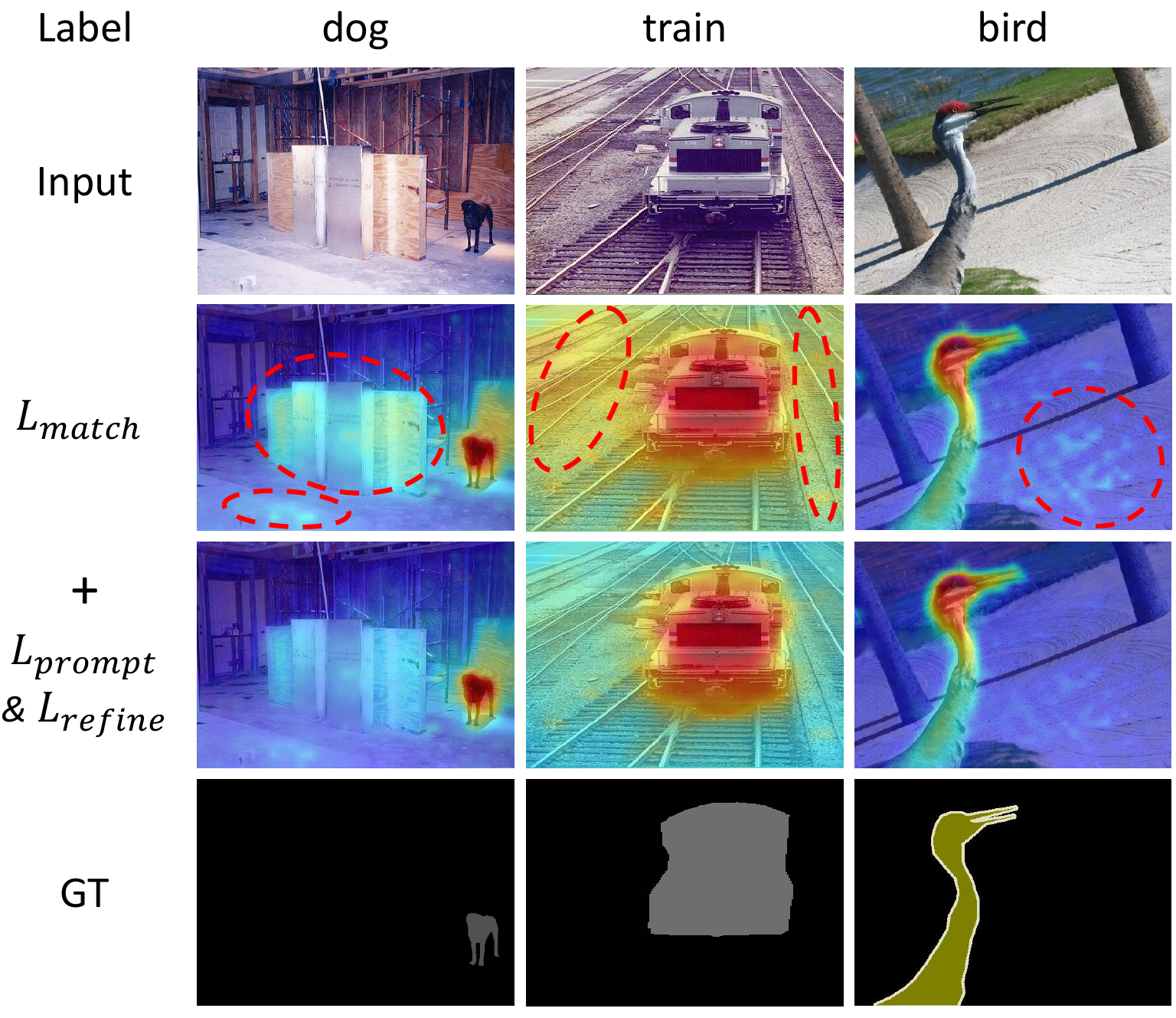}
  \caption{Qualitative ablation studies of loss functions. With both $L_{prompt}$ and $L_{refine}$ applied, the resulting CAMs are better aligned with the ground truth masks.
  }
  \label{figure:loss}
  \vspace{-4mm}
\end{figure}

To analyze the importance of the introduced loss functions, we conduct both quantitative and qualitative ablation studies, as shown in
Table~\ref{tab:loss} and Fig.~\ref{figure:loss}. In Table~\ref{tab:loss}, we see that applying only the matching loss $L_{match}$ would result in $67.6\%$ mIoU. If we directly add the refinement loss $L_{refine}$ without prompt learning, the performance would be similar since the prompts are randomly initialized and are not learned. When prompt learning is further considered, applying only the $L^T_{prompt}$ to repel the text labels may result in trivial solutions with little improvement. On the other hand, if only $L^I_{prompt}$ is enforced to align with the background images, the prompts are still likely to capture the semantics of the foreground object categories, resulting in $67.9\%$ mIoU. Finally, when $L^I_{prompt}$ and $L^T_{prompt}$ are jointly applied to learn the background regions while avoiding describing the foreground object categories, the mIoU would improve to $68.7\%$. Together with the qualitative results in Fig.~\ref{figure:loss}, we validate that our designed prompt learning and the proposed \nameshort~framework would prevent false activation of co-occurring backgrounds and therefore benefit segmentation in a weakly-supervised fashion.

\vspace{-2mm}
\section{Conclusion}
In this paper, we propose a \textit{\namebf~(\textbf{\nameshort})}~framework, which advances vision-language learning to achieve weakly-supervised semantic segmentation~(WSSS). In addition to exploiting the pre-trained CLIP model to perform \textit{\namea}, we further present \textit{\nameb}~and \textit{\namec}~in the proposed \textit{\nameshort}~framework to prevent false activation of image backgrounds. With no need to manually define background texts through prompt engineering, our learned prompts properly capture and suppress co-occurring backgrounds for each object category, resulting in precise activation maps for segmentation in a weakly-supervised fashion. Quantitative experiments on the segmentation benchmarks confirm the effectiveness of our proposed \textit{\nameshort}~framework, and visualization and ablation studies are conducted to demonstrate and verify the effectiveness of learned prompts.
Our method achieves competitive performance on the standard WSSS benchmarks, PASCAL VOC 2012 and MS COCO 2014, and shows compatibility with other WSSS methods.

\clearpage
\section{Appendix}
\begin{figure}[!h]
  \centering
  \includegraphics[width=\linewidth]{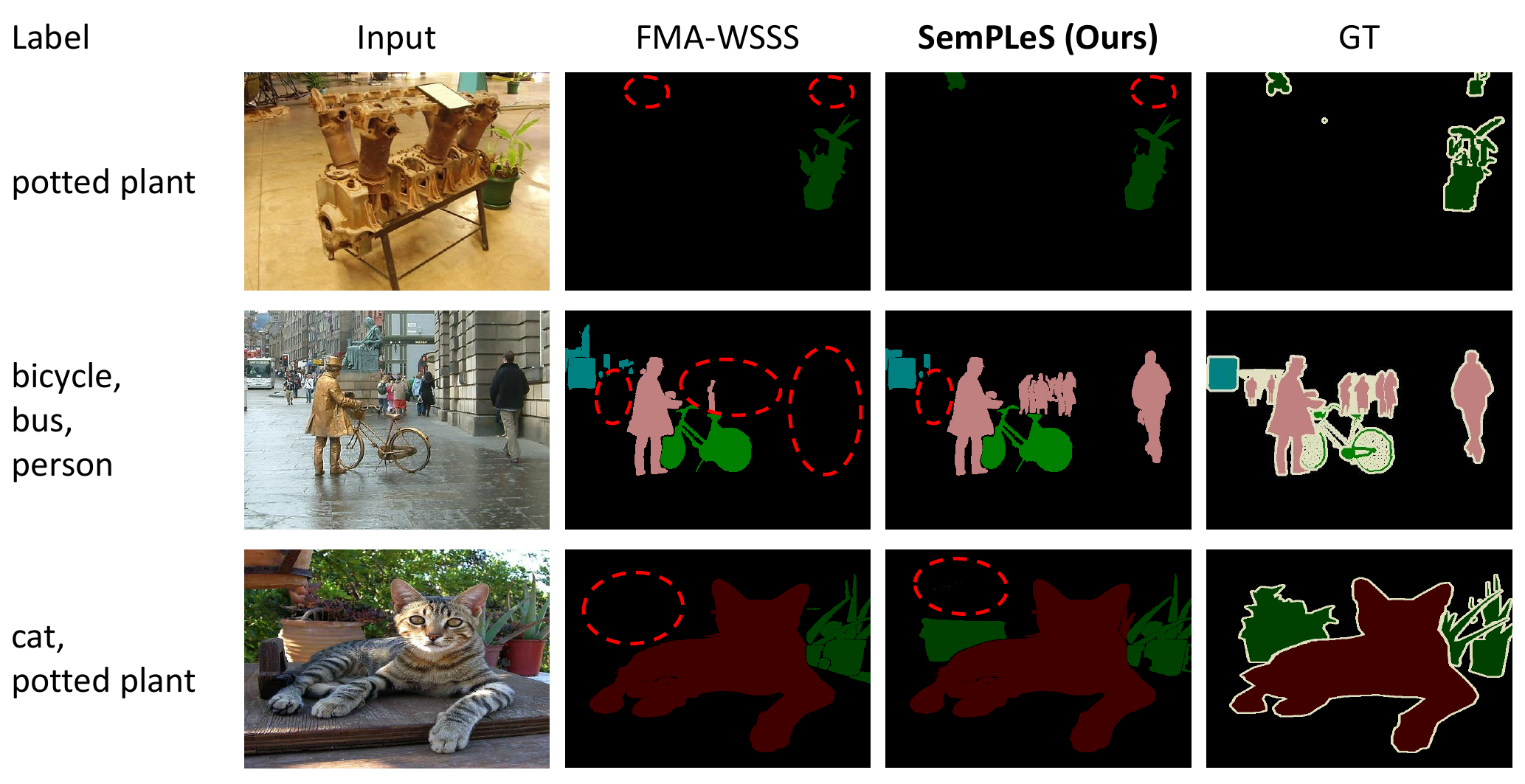}
  \caption{Failure case analysis. From the first to the third row, we show failure cases where the objects are partially visible, of small size, and visually similar to the surroundings, respectively.}
  \label{figure:failure}
   
\end{figure}

\subsection{Failure Case Analysis}

In Fig.~\ref{figure:failure}, we show several types of failure cases. In the first row, the potted plant in the top right corner is only partially visible in the image and thus is not easily recognized. In the second row, the two people in the top left corner are far away from the camera. The reduced size of such distant objects result in the failure of segmentation. As for the third row, the potted plant on the left is visually similar to its surroundings and therefore may confuse the models. Note that our \nameshort~still outperforms the previous SOTA (FMA-WSSS) in these challenging cases.

\subsection{Limitations and Potential Negative Impact}

\paragraph{\textbf{Limitations:}} In weakly-supervised semantic segmentation (WSSS), as only image-level labels are available for training, existing WSSS methods still struggle to segment the objects that are partially visible, of small size, or visually similar to the surroundings, as shown in Fig.~\ref{figure:failure}.

\paragraph{\textbf{Potential negative impact:}} No specific potential negative impact is identified in this work.

%%%%%%%%% REFERENCES
{\small
\bibliographystyle{ieee_fullname}
\bibliography{egbib}

\begin{thebibliography}{10}\itemsep=-1pt

\bibitem{ahn2019weakly}
Jiwoon Ahn, Sunghyun Cho, and Suha Kwak.
\newblock Weakly supervised learning of instance segmentation with inter-pixel relations.
\newblock In {\em CVPR}, 2019.

\bibitem{ahn2018learning}
Jiwoon Ahn and Suha Kwak.
\newblock Learning pixel-level semantic affinity with image-level supervision for weakly supervised semantic segmentation.
\newblock In {\em CVPR}, 2018.

\bibitem{bearman2016s}
Amy Bearman, Olga Russakovsky, Vittorio Ferrari, and Li Fei-Fei.
\newblock What’s the point: Semantic segmentation with point supervision.
\newblock In {\em ECCV}, 2016.

\bibitem{chang2020weakly}
Yu-Ting Chang, Qiaosong Wang, Wei-Chih Hung, Robinson Piramuthu, Yi-Hsuan Tsai, and Ming-Hsuan Yang.
\newblock Weakly-supervised semantic segmentation via sub-category exploration.
\newblock In {\em CVPR}, 2020.

\bibitem{chen2023fpr}
Liyi Chen, Chenyang Lei, Ruihuang Li, Shuai Li, Zhaoxiang Zhang, and Lei Zhang.
\newblock Fpr: False positive rectification for weakly supervised semantic segmentation.
\newblock In {\em ICCV}, 2023.

\bibitem{chen2017deeplab}
Liang-Chieh Chen, George Papandreou, Iasonas Kokkinos, Kevin Murphy, and Alan~L Yuille.
\newblock Deeplab: Semantic image segmentation with deep convolutional nets, atrous convolution, and fully connected crfs.
\newblock {\em TPAMI}, 2017.

\bibitem{chen2017rethinking}
Liang-Chieh Chen, George Papandreou, Florian Schroff, and Hartwig Adam.
\newblock Rethinking atrous convolution for semantic image segmentation.
\newblock {\em arXiv preprint arXiv:1706.05587}, 2017.

\bibitem{chen2018encoder}
Liang-Chieh Chen, Yukun Zhu, George Papandreou, Florian Schroff, and Hartwig Adam.
\newblock Encoder-decoder with atrous separable convolution for semantic image segmentation.
\newblock In {\em ECCV}, 2018.

\bibitem{chen2022self}
Qi Chen, Lingxiao Yang, Jian-Huang Lai, and Xiaohua Xie.
\newblock Self-supervised image-specific prototype exploration for weakly supervised semantic segmentation.
\newblock In {\em CVPR}, 2022.

\bibitem{chen2023segment}
Tianle Chen, Zheda Mai, Ruiwen Li, and Wei-lun Chao.
\newblock Segment anything model (sam) enhanced pseudo labels for weakly supervised semantic segmentation.
\newblock In {\em NeurIPS Workshop}, 2023.

\bibitem{chen2020uniter}
Yen-Chun Chen, Linjie Li, Licheng Yu, Ahmed El~Kholy, Faisal Ahmed, Zhe Gan, Yu Cheng, and Jingjing Liu.
\newblock Uniter: Universal image-text representation learning.
\newblock In {\em ECCV}, 2020.

\bibitem{chen2023extracting}
Zhaozheng Chen and Qianru Sun.
\newblock Extracting class activation maps from non-discriminative features as well.
\newblock In {\em CVPR}, 2023.

\bibitem{chen2023weakly}
Zhaozheng Chen and Qianru Sun.
\newblock Weakly-supervised semantic segmentation with image-level labels: from traditional models to foundation models.
\newblock {\em ACM Computing Surveys}, 2023.

\bibitem{chen2022class}
Zhaozheng Chen, Tan Wang, Xiongwei Wu, Xian-Sheng Hua, Hanwang Zhang, and Qianru Sun.
\newblock Class re-activation maps for weakly-supervised semantic segmentation.
\newblock In {\em CVPR}, 2022.

\bibitem{cheng2022masked}
Bowen Cheng, Ishan Misra, Alexander~G Schwing, Alexander Kirillov, and Rohit Girdhar.
\newblock Masked-attention mask transformer for universal image segmentation.
\newblock In {\em CVPR}, 2022.

\bibitem{cheng2023out}
Zesen Cheng, Pengchong Qiao, Kehan Li, Siheng Li, Pengxu Wei, Xiangyang Ji, Li Yuan, Chang Liu, and Jie Chen.
\newblock Out-of-candidate rectification for weakly supervised semantic segmentation.
\newblock In {\em CVPR}, 2023.

\bibitem{ding2022decoupling}
Jian Ding, Nan Xue, Gui-Song Xia, and Dengxin Dai.
\newblock Decoupling zero-shot semantic segmentation.
\newblock In {\em CVPR}, 2022.

\bibitem{dosovitskiy2021image}
Alexey Dosovitskiy, Lucas Beyer, Alexander Kolesnikov, Dirk Weissenborn, Xiaohua Zhai, Thomas Unterthiner, Mostafa Dehghani, Matthias Minderer, Georg Heigold, Sylvain Gelly, et~al.
\newblock An image is worth 16x16 words: Transformers for image recognition at scale.
\newblock In {\em ICLR}, 2021.

\bibitem{du2022weakly}
Ye Du, Zehua Fu, Qingjie Liu, and Yunhong Wang.
\newblock Weakly supervised semantic segmentation by pixel-to-prototype contrast.
\newblock In {\em CVPR}, 2022.

\bibitem{everingham2010pascal}
Mark Everingham, Luc Van~Gool, Christopher~KI Williams, John Winn, and Andrew Zisserman.
\newblock The pascal visual object classes (voc) challenge.
\newblock {\em IJCV}, 2010.

\bibitem{fan2020learning}
Junsong Fan, Zhaoxiang Zhang, Chunfeng Song, and Tieniu Tan.
\newblock Learning integral objects with intra-class discriminator for weakly-supervised semantic segmentation.
\newblock In {\em CVPR}, 2020.

\bibitem{ghiasi2022scaling}
Golnaz Ghiasi, Xiuye Gu, Yin Cui, and Tsung-Yi Lin.
\newblock Scaling open-vocabulary image segmentation with image-level labels.
\newblock In {\em ECCV}, 2022.

\bibitem{he2016deep}
Kaiming He, Xiangyu Zhang, Shaoqing Ren, and Jian Sun.
\newblock Deep residual learning for image recognition.
\newblock In {\em CVPR}, 2016.

\bibitem{he2023clip}
Wenbin He, Suphanut Jamonnak, Liang Gou, and Liu Ren.
\newblock Clip-{S}$^4$: Language-guided self-supervised semantic segmentation.
\newblock In {\em CVPR}, 2023.

\bibitem{jia2022visual}
Menglin Jia, Luming Tang, Bor-Chun Chen, Claire Cardie, Serge Belongie, Bharath Hariharan, and Ser-Nam Lim.
\newblock Visual prompt tuning.
\newblock In {\em ECCV}, 2022.

\bibitem{jiang2023segment}
Peng-Tao Jiang and Yuqi Yang.
\newblock Segment anything is a good pseudo-label generator for weakly supervised semantic segmentation.
\newblock {\em arXiv preprint arXiv:2305.01275}, 2023.

\bibitem{jiang2022l2g}
Peng-Tao Jiang, Yuqi Yang, Qibin Hou, and Yunchao Wei.
\newblock L2g: A simple local-to-global knowledge transfer framework for weakly supervised semantic segmentation.
\newblock In {\em CVPR}, 2022.

\bibitem{jo2023mars}
Sanghyun Jo, In-Jae Yu, and Kyungsu Kim.
\newblock Mars: Model-agnostic biased object removal without additional supervision for weakly-supervised semantic segmentation.
\newblock In {\em ICCV}, 2023.

\bibitem{khoreva2017simple}
Anna Khoreva, Rodrigo Benenson, Jan Hosang, Matthias Hein, and Bernt Schiele.
\newblock Simple does it: Weakly supervised instance and semantic segmentation.
\newblock In {\em CVPR}, 2017.

\bibitem{kirillov2023segment}
Alexander Kirillov, Eric Mintun, Nikhila Ravi, Hanzi Mao, Chloe Rolland, Laura Gustafson, Tete Xiao, Spencer Whitehead, Alexander~C Berg, Wan-Yen Lo, et~al.
\newblock Segment anything.
\newblock In {\em ICCV}, 2023.

\bibitem{krahenbuhl2011efficient}
Philipp Kr{\"a}henb{\"u}hl and Vladlen Koltun.
\newblock Efficient inference in fully connected crfs with gaussian edge potentials.
\newblock In {\em NeurIPS}, 2011.

\bibitem{kwon2024learning}
JuneHyoung Kwon, Eunju Lee, Yunsung Cho, and YoungBin Kim.
\newblock Learning to detour: Shortcut mitigating augmentation for weakly supervised semantic segmentation.
\newblock In {\em WACV}, 2024.

\bibitem{lee2022weakly}
Jungbeom Lee, Seong~Joon Oh, Sangdoo Yun, Junsuk Choe, Eunji Kim, and Sungroh Yoon.
\newblock Weakly supervised semantic segmentation using out-of-distribution data.
\newblock In {\em CVPR}, 2022.

\bibitem{li2022language}
Boyi Li, Kilian~Q Weinberger, Serge Belongie, Vladlen Koltun, and Ren{\'e} Ranftl.
\newblock Language-driven semantic segmentation.
\newblock In {\em ICLR}, 2022.

\bibitem{li2022towards}
Jing Li, Junsong Fan, and Zhaoxiang Zhang.
\newblock Towards noiseless object contours for weakly supervised semantic segmentation.
\newblock In {\em CVPR}, 2022.

\bibitem{li2022expansion}
Jinlong Li, Zequn Jie, Xu Wang, Xiaolin Wei, and Lin Ma.
\newblock Expansion and shrinkage of localization for weakly-supervised semantic segmentation.
\newblock In {\em NeurIPS}, 2022.

\bibitem{li2023blip}
Junnan Li, Dongxu Li, Silvio Savarese, and Steven Hoi.
\newblock Blip-2: Bootstrapping language-image pre-training with frozen image encoders and large language models.
\newblock In {\em ICML}, 2023.

\bibitem{liang2023open}
Feng Liang, Bichen Wu, Xiaoliang Dai, Kunpeng Li, Yinan Zhao, Hang Zhang, Peizhao Zhang, Peter Vajda, and Diana Marculescu.
\newblock Open-vocabulary semantic segmentation with mask-adapted clip.
\newblock In {\em CVPR}, 2023.

\bibitem{chen2015semantic}
Chen Liang-Chieh, George Papandreou, Iasonas Kokkinos, Kevin Murphy, et~al.
\newblock Semantic image segmentation with deep convolutional nets and fully connected crfs.
\newblock In {\em ICLR}, 2015.

\bibitem{lin2016scribblesup}
Di Lin, Jifeng Dai, Jiaya Jia, Kaiming He, and Jian Sun.
\newblock Scribblesup: Scribble-supervised convolutional networks for semantic segmentation.
\newblock In {\em CVPR}, 2016.

\bibitem{lin2014microsoft}
Tsung-Yi Lin, Michael Maire, Serge Belongie, James Hays, Pietro Perona, Deva Ramanan, Piotr Doll{\'a}r, and C~Lawrence Zitnick.
\newblock Microsoft coco: Common objects in context.
\newblock In {\em ECCV}, 2014.

\bibitem{lin2023clip}
Yuqi Lin, Minghao Chen, Wenxiao Wang, Boxi Wu, Ke Li, Binbin Lin, Haifeng Liu, and Xiaofei He.
\newblock Clip is also an efficient segmenter: A text-driven approach for weakly supervised semantic segmentation.
\newblock In {\em CVPR}, 2023.

\bibitem{liu2023pre}
Pengfei Liu, Weizhe Yuan, Jinlan Fu, Zhengbao Jiang, Hiroaki Hayashi, and Graham Neubig.
\newblock Pre-train, prompt, and predict: A systematic survey of prompting methods in natural language processing.
\newblock {\em ACM Computing Surveys}, 2023.

\bibitem{liu2023grounding}
Shilong Liu, Zhaoyang Zeng, Tianhe Ren, Feng Li, Hao Zhang, Jie Yang, Chunyuan Li, Jianwei Yang, Hang Su, Jun Zhu, et~al.
\newblock Grounding dino: Marrying dino with grounded pre-training for open-set object detection.
\newblock In {\em ECCV}, 2024.

\bibitem{long2015fully}
Jonathan Long, Evan Shelhamer, and Trevor Darrell.
\newblock Fully convolutional networks for semantic segmentation.
\newblock In {\em CVPR}, 2015.

\bibitem{lu2019vilbert}
Jiasen Lu, Dhruv Batra, Devi Parikh, and Stefan Lee.
\newblock Vilbert: Pretraining task-agnostic visiolinguistic representations for vision-and-language tasks.
\newblock In {\em NeurIPS}, 2019.

\bibitem{luddecke2022image}
Timo L{\"u}ddecke and Alexander Ecker.
\newblock Image segmentation using text and image prompts.
\newblock In {\em CVPR}, 2022.

\bibitem{meyer2017improved}
Benjamin~J Meyer and Tom Drummond.
\newblock Improved semantic segmentation for robotic applications with hierarchical conditional random fields.
\newblock In {\em ICRA}, 2017.

\bibitem{murugesan2024prompting}
Balamurali Murugesan, Rukhshanda Hussain, Rajarshi Bhattacharya, Ismail Ben~Ayed, and Jose Dolz.
\newblock Prompting classes: Exploring the power of prompt class learning in weakly supervised semantic segmentation.
\newblock In {\em WACV}, 2024.

\bibitem{peng2023usage}
Zelin Peng, Guanchun Wang, Lingxi Xie, Dongsheng Jiang, Wei Shen, and Qi Tian.
\newblock Usage: A unified seed area generation paradigm for weakly supervised semantic segmentation.
\newblock In {\em ICCV}, 2023.

\bibitem{radford2021learning}
Alec Radford, Jong~Wook Kim, Chris Hallacy, Aditya Ramesh, Gabriel Goh, Sandhini Agarwal, Girish Sastry, Amanda Askell, Pamela Mishkin, Jack Clark, et~al.
\newblock Learning transferable visual models from natural language supervision.
\newblock In {\em ICML}, 2021.

\bibitem{rao2022denseclip}
Yongming Rao, Wenliang Zhao, Guangyi Chen, Yansong Tang, Zheng Zhu, Guan Huang, Jie Zhou, and Jiwen Lu.
\newblock Denseclip: Language-guided dense prediction with context-aware prompting.
\newblock In {\em CVPR}, 2022.

\bibitem{ronneberger2015u}
Olaf Ronneberger, Philipp Fischer, and Thomas Brox.
\newblock U-net: Convolutional networks for biomedical image segmentation.
\newblock In {\em MICCAI}, 2015.

\bibitem{rossetti2022max}
Simone Rossetti, Damiano Zappia, Marta Sanzari, Marco Schaerf, and Fiora Pirri.
\newblock Max pooling with vision transformers reconciles class and shape in weakly supervised semantic segmentation.
\newblock In {\em ECCV}, 2022.

\bibitem{ru2022learning}
Lixiang Ru, Yibing Zhan, Baosheng Yu, and Bo Du.
\newblock Learning affinity from attention: end-to-end weakly-supervised semantic segmentation with transformers.
\newblock In {\em CVPR}, 2022.

\bibitem{ru2023token}
Lixiang Ru, Heliang Zheng, Yibing Zhan, and Bo Du.
\newblock Token contrast for weakly-supervised semantic segmentation.
\newblock In {\em CVPR}, 2023.

\bibitem{selvaraju2017grad}
Ramprasaath~R Selvaraju, Michael Cogswell, Abhishek Das, Ramakrishna Vedantam, Devi Parikh, and Dhruv Batra.
\newblock Grad-cam: Visual explanations from deep networks via gradient-based localization.
\newblock In {\em ICCV}, 2017.

\bibitem{shin2022reco}
Gyungin Shin, Weidi Xie, and Samuel Albanie.
\newblock Reco: Retrieve and co-segment for zero-shot transfer.
\newblock In {\em NeurIPS}, 2022.

\bibitem{sun2023alternative}
Weixuan Sun, Zheyuan Liu, Yanhao Zhang, Yiran Zhong, and Nick Barnes.
\newblock An alternative to wsss? an empirical study of the segment anything model (sam) on weakly-supervised semantic segmentation problems.
\newblock {\em arXiv preprint arXiv:2305.01586}, 2023.

\bibitem{wang2023treating}
Changwei Wang, Rongtao Xu, Shibiao Xu, Weiliang Meng, and Xiaopeng Zhang.
\newblock Treating pseudo-labels generation as image matting for weakly supervised semantic segmentation.
\newblock In {\em ICCV}, 2023.

\bibitem{wang2020self}
Yude Wang, Jie Zhang, Meina Kan, Shiguang Shan, and Xilin Chen.
\newblock Self-supervised equivariant attention mechanism for weakly supervised semantic segmentation.
\newblock In {\em CVPR}, 2020.

\bibitem{wu2024masked}
Fangwen Wu, Jingxuan He, Lechao Cheng, Yufei Yin, Yanbin Hao, and Gang Huang.
\newblock Masked collaborative contrast for weakly supervised semantic segmentation.
\newblock In {\em WACV}, 2024.

\bibitem{wu2022adaptive}
Tong Wu, Guangyu Gao, Junshi Huang, Xiaolin Wei, Xiaoming Wei, and Chi~Harold Liu.
\newblock Adaptive spatial-bce loss for weakly supervised semantic segmentation.
\newblock In {\em ECCV}, 2022.

\bibitem{wu2024dupl}
Yuanchen Wu, Xichen Ye, Kequan Yang, Jide Li, and Xiaoqiang Li.
\newblock Dupl: Dual student with trustworthy progressive learning for robust weakly supervised semantic segmentation.
\newblock In {\em CVPR}, 2024.

\bibitem{xie2022clims}
Jinheng Xie, Xianxu Hou, Kai Ye, and Linlin Shen.
\newblock Clims: cross language image matching for weakly supervised semantic segmentation.
\newblock In {\em CVPR}, 2022.

\bibitem{xie2022c2am}
Jinheng Xie, Jianfeng Xiang, Junliang Chen, Xianxu Hou, Xiaodong Zhao, and Linlin Shen.
\newblock C2am: contrastive learning of class-agnostic activation map for weakly supervised object localization and semantic segmentation.
\newblock In {\em CVPR}, 2022.

\bibitem{xu2023open}
Jiarui Xu, Sifei Liu, Arash Vahdat, Wonmin Byeon, Xiaolong Wang, and Shalini De~Mello.
\newblock Open-vocabulary panoptic segmentation with text-to-image diffusion models.
\newblock In {\em CVPR}, 2023.

\bibitem{xu2022multi}
Lian Xu, Wanli Ouyang, Mohammed Bennamoun, Farid Boussaid, and Dan Xu.
\newblock Multi-class token transformer for weakly supervised semantic segmentation.
\newblock In {\em CVPR}, 2022.

\bibitem{xu2023mmcst}
Lian Xu, Wanli Ouyang, Mohammed Bennamoun, Farid Boussaid, and Dan Xu.
\newblock Learning multi-modal class-specific tokens for weakly supervised dense object localization.
\newblock In {\em CVPR}, 2023.

\bibitem{xu2023side}
Mengde Xu, Zheng Zhang, Fangyun Wei, Han Hu, and Xiang Bai.
\newblock Side adapter network for open-vocabulary semantic segmentation.
\newblock In {\em CVPR}, 2023.

\bibitem{xu2022simple}
Mengde Xu, Zheng Zhang, Fangyun Wei, Yutong Lin, Yue Cao, Han Hu, and Xiang Bai.
\newblock A simple baseline for open-vocabulary semantic segmentation with pre-trained vision-language model.
\newblock In {\em ECCV}, 2022.

\bibitem{yang2024foundation}
Xiaobo Yang and Xiaojin Gong.
\newblock Foundation model assisted weakly supervised semantic segmentation.
\newblock In {\em WACV}, 2024.

\bibitem{yoon2022adversarial}
Sung-Hoon Yoon, Hyeokjun Kweon, Jegyeong Cho, Shinjeong Kim, and Kuk-Jin Yoon.
\newblock Adversarial erasing framework via triplet with gated pyramid pooling layer for weakly supervised semantic segmentation.
\newblock In {\em ECCV}, 2022.

\bibitem{zagoruyko2016wide}
Sergey Zagoruyko and Nikos Komodakis.
\newblock Wide residual networks.
\newblock In {\em BMVC}, 2016.

\bibitem{zendel2022unifying}
Oliver Zendel, Matthias Sch{\"o}rghuber, Bernhard Rainer, Markus Murschitz, and Csaba Beleznai.
\newblock Unifying panoptic segmentation for autonomous driving.
\newblock In {\em CVPR}, 2022.

\bibitem{zhang2023recognize}
Youcai Zhang, Xinyu Huang, Jinyu Ma, Zhaoyang Li, Zhaochuan Luo, Yanchun Xie, Yuzhuo Qin, Tong Luo, Yaqian Li, Shilong Liu, et~al.
\newblock Recognize anything: A strong image tagging model.
\newblock In {\em CVPR}, 2024.

\bibitem{zhao2017pyramid}
Hengshuang Zhao, Jianping Shi, Xiaojuan Qi, Xiaogang Wang, and Jiaya Jia.
\newblock Pyramid scene parsing network.
\newblock In {\em CVPR}, 2017.

\bibitem{zhao2018psanet}
Hengshuang Zhao, Yi Zhang, Shu Liu, Jianping Shi, Chen~Change Loy, Dahua Lin, and Jiaya Jia.
\newblock Psanet: Point-wise spatial attention network for scene parsing.
\newblock In {\em ECCV}, 2018.

\bibitem{zhao2024sfc}
Xinqiao Zhao, Feilong Tang, Xiaoyang Wang, and Jimin Xiao.
\newblock Sfc: Shared feature calibration in weakly supervised semantic segmentation.
\newblock In {\em AAAI}, 2024.

\bibitem{zhou2016learning}
Bolei Zhou, Aditya Khosla, Agata Lapedriza, Aude Oliva, and Antonio Torralba.
\newblock Learning deep features for discriminative localization.
\newblock In {\em CVPR}, 2016.

\bibitem{zhou2022extract}
Chong Zhou, Chen~Change Loy, and Bo Dai.
\newblock Extract free dense labels from clip.
\newblock In {\em ECCV}, 2022.

\bibitem{zhou2022conditional}
Kaiyang Zhou, Jingkang Yang, Chen~Change Loy, and Ziwei Liu.
\newblock Conditional prompt learning for vision-language models.
\newblock In {\em CVPR}, 2022.

\bibitem{zhou2022learning}
Kaiyang Zhou, Jingkang Yang, Chen~Change Loy, and Ziwei Liu.
\newblock Learning to prompt for vision-language models.
\newblock {\em IJCV}, 2022.

\bibitem{zhu2023weaktr}
Lianghui Zhu, Yingyue Li, Jieming Fang, Yan Liu, Hao Xin, Wenyu Liu, and Xinggang Wang.
\newblock Weaktr: Exploring plain vision transformer for weakly-supervised semantic segmentation.
\newblock {\em arXiv preprint arXiv:2304.01184}, 2023.

\end{thebibliography}
}

\end{document}